\documentclass[letterpaper, 10 pt, conference]{style}
\IEEEoverridecommandlockouts
\usepackage{cite}
\usepackage{graphicx}

\usepackage{svg}
\usepackage{amsmath,amssymb,amsfonts}
\usepackage[
  bookmarks=false,
  colorlinks=true,
  linkcolor=blue,
  citecolor=blue,
  urlcolor=blue
]{hyperref}\usepackage{algorithmic}
\usepackage{textcomp}
\usepackage{booktabs}
\usepackage{amsmath} 
\usepackage{tabularx} 
\usepackage{xcolor}
\def\BibTeX{{\rm B\kern-.05em{\sc i\kern-.025em b}\kern-.08em
    T\kern-.1667em\lower.7ex\hbox{E}\kern-.125emX}}

\begin{document}

\title{\LARGE \bf Capacity-Aware Deep Learning for Generalizable Traffic Volume Estimation Across Links and Cities}

\author{Léo Hein$^{1,2}$, Giovanni De Nunzio$^{1}$, Aurélie Pirayre$^{1}$, Laurent Najman$^{2,3}$
\thanks{$^{1}$IFP Energies nouvelles, 1 et 4 avenue de Bois-Préau, 92852 Rueil-Malmaison, France,
        {\footnotesize \{leo.hein, aurelie.chataignon, giovanni.de-nunzio\}@ifpen.fr}}%
 \thanks{$^{2}$Université Gustave Eiffel, CNRS, LIGM, Marne-la-Vallée, 77454, France, {\footnotesize laurent.najman@esiee.fr}}%
 \thanks{$^{3}$Khalifa University, Math Department, Abu Dhabi, UAE}}%

\maketitle
\begin{abstract}
Network-wide traffic volume estimation typically relies on propagating measurements from fixed sensors, making performance highly dependent on sensor density and limiting deployment in sparsely instrumented networks. We propose a link-level learning framework that estimates hourly traffic volumes from widely available territorial data only, including probe speed profiles, road and topological descriptors, along with weather observations. A supervised local mapping is learned from sparse sensor measurements and evaluated under two generalization settings: intra-network (unseen links within the training network) and inter-network (unseen city). This formulation frames traffic volume estimation as a spatial out-of-distribution generalization problem under sparse supervision. To enhance spatial robustness, we introduce a capacity-aware formulation that models volume as the product of a link-specific structural capacity and an hourly regime-aware utilization ratio, embedding traffic-theoretic constraints directly into the learning process. Extensive experiments in both generalization settings demonstrate that the proposed structural constraints consistently outperform a state-of-the-art baseline under spatial distribution shift.
\end{abstract}


\section{Introduction}

Estimating accurate traffic volumes across an urban road network is critical for urban planning, traffic management, and environmental monitoring, including air quality assessment \cite{air quality}. Although traffic counts can be precisely collected using fixed sensors, large-scale deployment and maintenance remain costly. Spatial propagation methods are commonly used to extrapolate sparse counts to uninstrumented segments\cite{b14,b15,b16}; however, their effectiveness relies on adequate sensor coverage, which limits their applicability in sparsely instrumented or entirely unsensed cities. In contrast, link speeds (e.g., floating-car data), static road and topological attributes, and meteorological observations are widely available across most territories. This motivates approaches that estimate traffic volumes solely from broadly accessible data sources, without requiring traffic counts at inference time.

Many classical approaches rely on parametric fundamental diagrams which, although physically grounded, impose fixed functional forms that may introduce structural bias and limit adaptability across road types and operating conditions \cite{fd1}, as most formulations depend solely on speed-density or speed-flow relationships without incorporating contextual factors.

Supervised machine learning models have therefore been proposed to learn speed-flow mappings, enabling more flexible functional forms and simplifying calibration by relying directly on observed measurements rather than predefined parametric assumptions \cite{ml1,ml2}. In this context, temporal deep architectures, including convolutional, recurrent, and Transformer-based models, have proven effective at capturing intraday dynamics in the speed-volume translation paradigm. 

In addition, spatio-temporal graph neural networks (STGNNs) incorporate road connectivity through message passing and achieve strong performance in many traffic prediction tasks \cite{stgnn}. However, most formulations are transductive and tied to a fixed network topology, limiting transfer to new cities without retraining \cite{stgnn, dcrnn}. To improve spatial generalization, recent works adopt inductive link-level formulations that rely on sensor-free link attributes and contextual features as transferable inputs \cite{hammoumi2025}. Some others further improve the exploitation of local topological information through neighborhood-based graph neural networks \cite{hein2025}. Yet under sparse supervision, learning a robust transferable mapping remains challenging, as high-capacity models risk overfitting the labeled subset.

Cross-city deployment also relates to domain adaptation and transfer learning under distribution shift. However, most existing approaches assume access to target-domain supervision, which is used for fine-tuning, domain-specific recalibration, adversarial feature alignment, or graph-dependent representation adaptation \cite{transfer1, transfer2, transfer3, transfer4}.

In this work, we address a spatial out-of-distribution (OOD) setting where supervision is available only for a sparse subset of links in a source network, and inference must generalize to unseen links and new cities without additional calibration. Under such weak supervision, purely increasing model complexity is unlikely to improve generalization \cite{generalization}. Instead, we argue that the learning process must be guided by structural constraints that encode physically meaningful traffic behavior.

We adopt an inductive, link-level formulation that estimates hourly traffic volumes from same-day speed profiles, temporal context, meteorological variables, and sensor-free road descriptors augmented with lightweight topological cues. To improve spatial generalization under limited data and sparse supervision, we embed traffic-theoretic structure directly into the mapping through two main contributions:

\begin{itemize}
    \item \textbf{Capacity-aware bounded prediction.}  
    We decompose traffic volume into a non-negative link-specific capacity and a bounded utilization ratio. This factorization enforces physical consistency while enhancing robustness to spatial distribution shifts.

    \item \textbf{Regime-aware utilization modeling.}  
    We capture both pre-capacity and post-capacity regimes using a softly gated mixture mechanism, enabling the model to represent non-monotonic traffic dynamics within a unified bounded framework.
\end{itemize}

\section{Problem Definition} We formulate the problem as link-level traffic volume estimation under a spatial out-of-distribution (OOD) generalization setting. The objective is to learn a transformation that maps widely available, volume-related data, such as probe vehicle speeds, static road infrastructure, network topology descriptors, and weather conditions, to a daily hourly traffic volume profile. The learned mapping is designed to generalize to unseen links within the training network and to new networks drawn from potentially different spatial distributions. This setting induces distribution shifts across link attributes, traffic regimes, and network topology. 

Let $\mathcal{N}$ denote a road network composed of a set of links $\mathcal{L}$ and let $\mathcal{D}$ denote the set of available days. For each link $\ell \in \mathcal{L}$ and day $d \in \mathcal{D}$, we observe input features \[ X_{\ell,d} = (x^{static}_\ell, x^{temp}_{\ell,d}), \] where $x^{static}_\ell$ denotes a vector of static link descriptors (road attributes and local topology features) and $x^{temp}_{\ell,d}$ denotes a vector of temporal inputs (daily speed profiles augmented with calendar and weather variables). 

Consider now a source network $\mathcal{N}^A$ with link set $\mathcal{L}^A$ and let $\mathcal{L}_{obs}^A \subset \mathcal{L}^A$ denote the subset of instrumented links equipped with traffic sensors. For each $\ell \in \mathcal{L}_{obs}^A$ and day $d$, the corresponding hourly traffic volume profile $q_{\ell,d}$ is observed. The goal is to learn a mapping \[ f_\theta : X_{\ell,d} \rightarrow \hat{q}_{\ell,d}, \] where $\theta$ denotes learnable parameters and $\hat{q}_{\ell,d}$ the estimated volume profile for link $\ell$ and day $d$. The parameters $\theta$ are estimated using supervision available only on $\mathcal{L}_{obs}^A$, while the learned mapping is intended to generalize to all links, including those without supervision and those belonging to unseen networks. 

The learned model is evaluated under two distinct generalization regimes: \begin{itemize} \item \textbf{Intra-network generalization:} performance on unseen links within the same network, reflecting generalization to unseen links under partial network supervision. \item \textbf{Inter-network generalization:} performance on a separate and unseen network $\mathcal{N}^B$, whose link attributes, topology, and traffic dynamics may follow a different distribution from those of $\mathcal{N}^A$. \end{itemize}

\section{Methodology}

\subsection{Architecture Overview}

\begin{figure*}[t]
\centering
\includegraphics[
    width=\textwidth,
    trim=0.5cm 0.3cm 2cm 0.2cm,
    clip
]{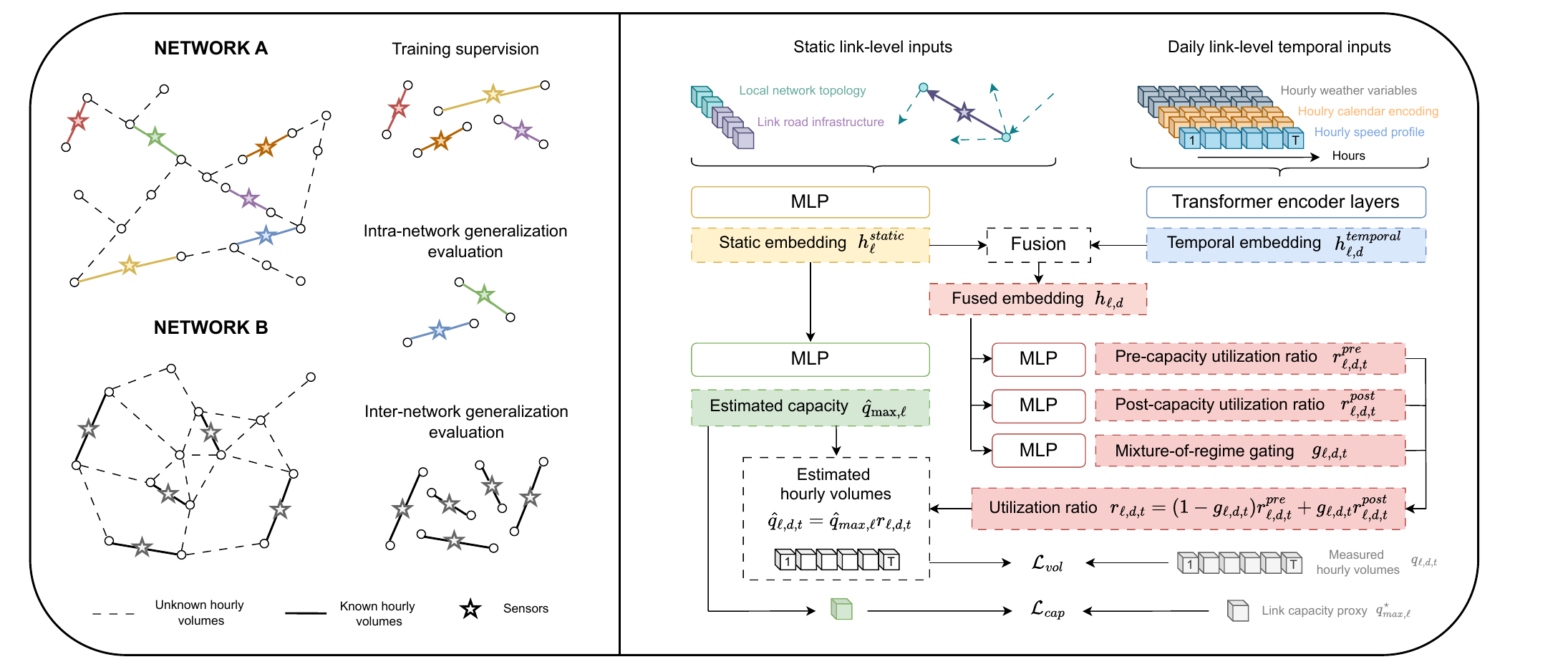}
\caption{Overview of the proposed framework for our generalizable traffic volume estimation approach.
(Left) Multi-network generalization setup: the model is trained on a source network with partially observed volume profiles and evaluated under both intra-network and inter-network generalization. The graph is directed, and sensor directions are omitted for visual clarity.
(Right) Model architecture: static link-level features are encoded with an MLP, while daily temporal inputs are encoded with Transformer layers. Both embeddings are fused to estimate daily utilization ratio through a mixture-of-regimes mechanism, while the static embedding is fed to an MLP to estimate the link capacity. The estimated volume profile is then computed through a capacity-utilization ratio formula. The model is trained with capacity and volume supervision.}
\label{fig:main}
\end{figure*}

We propose a hybrid multimodal architecture for inductive link-level traffic volume estimation that maps heterogeneous descriptors to a full-day hourly volume profile without requiring traffic counts at inference time. As illustrated in Figure~\ref{fig:main}, the model consists of two encoding branches handling static and temporal inputs. The static branch processes time-invariant link descriptors (road infrastructure and local topology) through a Multilayer Perceptron (MLP). The temporal branch uses Transformer encoders to capture intra-day traffic dynamics from speed profiles enriched with calendar and weather variables. The resulting embeddings are fused to parameterize a capacity-aware bounded output, where hourly volume is modeled as the product of an estimated link-specific capacity and a regime-aware utilization ratio. The capacity is inferred from static descriptors, while the utilization is derived from the fused representation via gating. 

By decoupling structural capacity from demand dynamics and operating independently at the link level without relying on a fixed global graph structure, the proposed design embeds traffic-theoretic structure and ensures inductive generalization to unseen links and networks under spatial distribution shift.

\subsection{Static Branch}

\textbf{Static descriptors.} 
For each link $\ell$ in $\mathcal{L}_{obs}^A$, the static feature vector $x^{static}_\ell \in \mathbb{R}^{C_{static}}$ gathers time-invariant attributes describing its structural characteristics, where $C_{static}$ is the number of static features. These include standard road infrastructure variables such as the number of lanes as well as lightweight topology descriptors capturing local road connectivity within the directed network. Together, these features encode intrinsic structural capacity and typical traffic intensity patterns associated with the link.

\textbf{Static encoding.}
The static encoder maps $x^{static}_\ell$ into a latent representation $z^{(0)}_\ell \in \mathbb{R}^{C_{sh}}$, with $C_{sh}$ being the static branch hidden dimension, through a learnable projection. Then, an MLP composed of $L_{static}$ fully connected layers operating in $\mathbb{R}^{C_{sh}}$ is applied. For $k = 1, \dots, L_{static}$, the transformation is defined as
\[
z^{(k)}_\ell =
\mathrm{Dropout}\big(
\mathrm{ReLU}(
\mathrm{LayerNorm}(W_k z^{(k-1)}_\ell)
)\big),
\]
where $W_k \in \mathbb{R}^{C_{sh} \times C_{sh}}$ are learnable weight matrices.

The final embedding
\[
h^{static}_\ell = z^{(L_{static})}_\ell \in \mathbb{R}^{C_{sh}}
\]
summarizes the link’s structural properties and provides the basis for estimating its nominal traffic capacity within the capacity-aware formulation.

\subsection{Temporal Branch}

\textbf{Temporal descriptors.}  
For each link $\ell$ in $\mathcal{L}_{obs}^A$ and day $d$ in $\mathcal{D}$, the temporal input $x^{temp}_{\ell,d} \in \mathbb{R}^{T \times C_{temp}}$ represents an hourly feature sequence, where $T$ is $24$ and $C_{temp}$ is the number of temporal features. Each hourly vector concatenates speed measurements, weather variables, and calendar encodings:
\[
\forall t \in \{1,\dots,T\}, \quad
x^{\text{temp}}_{\ell,d,t}
=
\left[
x^{\text{speed}}_{\ell,d,t}
\;\|\;
x^{\text{weather}}_{\ell,d,t}
\;\|\;
x^{\text{cal}}_{d,t}
\right].
\]
\textbf{Temporal encoding.}
The temporal encoder projects each hourly feature vector into a latent space $\mathbb{R}^{C_{th}}$, with $C_{th}$ being the temporal branch hidden dimension, and processes the resulting sequence using $L_{temp}$ stacked non-causal Transformer encoder layers. The non-causal design allows information from any hour to inform representations at other time steps, capturing global intra-day traffic dependencies.

Let $y^{(0)}_{\ell,d} \in \mathbb{R}^{T \times C_{th}}$ denote the projected sequence. Each Transformer layer applies multi-head self-attention with residual connections and a position-wise feed-forward network:
\[
y^{(k)}_{\ell,d} =
\mathrm{TransformerLayer}\!\left(
y^{(k-1)}_{\ell,d}
\right),
\quad k=1,\dots,L_{temp}.
\]

The final temporal embedding
\[
h^{temp}_{\ell,d}
=
y^{(L_{temp})}_{\ell,d}
\in
\mathbb{R}^{T \times C_{th}}
\]
encodes the full intra-day evolution of traffic conditions and contextual factors for link $\ell$ on day $d$, and provides the time-resolved features used in downstream utilization modeling.

\subsection{Multimodal Fusion}

The static embedding $h^{static}_\ell \in \mathbb{R}^{C_{sh}}$ encodes time-invariant link attributes, while the temporal embedding 
$h^{temp}_{\ell,d} \in \mathbb{R}^{T \times C_{th}}$ captures intra-day traffic dynamics. To integrate structural and temporal information at each time step, the static embedding is broadcast across the temporal dimension:
\[
\tilde{h}^{static}_{\ell}
=
\mathrm{Repeat}(h^{static}_\ell, T)
\in
\mathbb{R}^{T \times C_{sh}}.
\]

The fused representation is obtained by concatenation followed by layer normalization:
\[
h_{\ell,d}
=
\mathrm{LayerNorm}
\Big(
[\tilde{h}^{static}_{\ell} \;\|\; h^{temp}_{\ell,d}]
\Big)
\in
\mathbb{R}^{T \times (C_{sh} + C_{th})}.
\]

It provides a time-resolved multimodal representation combining structural capacity cues and dynamic traffic conditions, and serves as input to the downstream predictive heads.

\subsection{Physically Constrained Volume Estimation Branch}

\textbf{Capacity-Aware Bounded Output.} 
Predicting absolute hourly volumes without structural constraints forces the model to jointly learn structural scale and temporal usage patterns, which may yield physically implausible solutions and unstable behavior under distribution shifts. Since link volume cannot exceed its physical capacity, we factorize the estimation of the volume $\hat{q}_{\ell,d,t}$ into a non-negative structural capacity $\hat{q}_{\max,\ell} \in \mathbb{R}_+$ and a bounded utilization ratio $r_{\ell,d,t} \in [0,1]$:
\[
\hat{q}_{\ell,d,t}
=
\hat{q}_{\max,\ell}\, r_{\ell,d,t}.
\]
This formulation enforces the upper bound $\hat{q}_{\ell,d,t} \leq \hat{q}_{\max,\ell}$, guarantees non-negative outputs, and removes scale ambiguity by separating structural capacity from utilization, thereby restricting the model to physically admissible solutions and promoting more stable generalization under distribution shifts.

Capacity is estimated exclusively from the static embedding:
\[
\hat{q}_{\max,\ell}
=
\mathrm{softplus}\!\left(\psi_{cap}(h^{static}_\ell)\right),
\]
where $\psi_{cap}:\mathbb{R}^{C_{sh}}\rightarrow\mathbb{R}$ is an MLP and $\mathrm{softplus}(\cdot)$ enforces positivity while preserving differentiability. During training, this head is supervised using an empirical capacity proxy $q^{\star}_{\max,\ell}$ computed from historical volume observations over the available observation horizon $\mathcal{D}$. Specifically, for each instrumented link $\ell \in \mathcal{L}_{obs}^A$, we define $q^{\star}_{\max,\ell}$ the 99th percentile of all observed hourly volumes across the available days and time steps:
\[
q^{\star}_{\max,\ell}
=
\mathrm{Quantile}_{0.99}
\big(
\{ q_{\ell,d,t} \}_{d \in \mathcal{D},\, t=1,\dots,T}
\big).
\]
Importantly, we additionally verify, using a predefined statistical criterion, that the resulting quantile lies within the upper plateau of the historical hourly flow distribution of the link, supporting consistency with near-capacity operating conditions. The associated supervision loss is:
\[
\mathcal{L}_{cap}
=
\frac{1}{|\mathcal{E}|}
\sum_{\ell}
\left(
\hat{q}_{\max,\ell}
-
q^{\star}_{\max,\ell}
\right)^2,
\]
where $\mathcal{E}$ denotes the set of instrumented links. This capacity supervision acts as an auxiliary structural constraint that ties predicted volumes to link-specific capacity levels determined by local road attributes and topology. This alignment makes predictions more consistent with traffic flow theory and more robust to spatial transfer. 

\textbf{Utilization Heads and Regime Gating.}
Although the bounded formulation $\hat{q}_{\ell,d,t} = \hat{q}_{\max,\ell} r_{\ell,d,t}$ ensures physical consistency, a single utilization ratio cannot capture the non-monotonic demand-flow relationship. Since volume increases up to capacity and decreases under congestion, identical normalized levels $r_{\ell,d,t}$ may arise from either demand-limited or congestion-limited regimes, leading to ambiguity in regime interpretation.

To account for this regime heterogeneity while preserving boundedness, we adopt a soft mixture-of-regimes parameterization. At each hour, three task-specific MLP heads $\psi_{\cdot} : \mathbb{R}^{C_{sh}+C_{th}} \rightarrow \mathbb{R}$ are applied to the fused representation $h_{\ell,d,t}$. Two heads produce regime-specific logits:
\[
r^{pre}_{\ell,d,t}
=
\sigma\!\left(\psi_{pre}(h_{\ell,d,t})\right),
\qquad
r^{post}_{\ell,d,t}
=
\sigma\!\left(\psi_{post}(h_{\ell,d,t})\right),
\]
where $\sigma(\cdot)$ denotes the sigmoid function, 
$r^{pre}$ models utilization in demand-limited conditions and 
$r^{post}$ captures effective utilization under congestion. A third head produces a soft regime weight
\[
g_{\ell,d,t}
=
\sigma(\psi_{gate}(h_{\ell,d,t})),
\]
which blends both regimes:
\[
r_{\ell,d,t}
=
(1-g_{\ell,d,t})\, r^{pre}_{\ell,d,t}
+
g_{\ell,d,t}\, r^{post}_{\ell,d,t}.
\]

The gating term $g_{\ell,d,t}$ allows the model to adjust the regime contribution at an hourly resolution, as demand-limited and congestion-limited conditions may alternate within the same link over the course of an hour. Rather than imposing a hard regime switch, the soft mixture captures gradual transitions and regime uncertainty.

\section{Experiments}

\subsection{Dataset Construction}

We construct two urban traffic datasets (Lyon and Nantes) using a unified processing pipeline to ensure feature consistency across networks. Each dataset combines traffic counts, probe-based speeds, static link attributes, topology-derived features, meteorological variables, and calendar encodings defined on the directed road graph. The two cities exhibit different network sizes, sensor densities, and traffic intensity distributions, inducing non-trivial cross-network shifts.

\textbf{Static features.}
Road topology and link attributes are extracted from HERE Maps to reconstruct the directed primal graph (intersections as nodes, road segments as edges). For each link from upstream node $u$ to downstream node $v$, we compute the in- and out-degrees of both nodes as lightweight connectivity indicators capturing local intersection complexity, while avoiding higher-order graph metrics prone to overfitting under sparse supervision. Additional infrastructural attributes include functional class (1--5), number of lanes, speed limit, link length, free-flow speed, curvature, and slope.

\textbf{Temporal features.}
Hourly probe-based speeds are extracted from HERE Maps for all directed links. Each timestamp is augmented with calendar covariates (hour, weekday, day, month, public holidays and school vacation periods indicator). Hourly weather variables (precipitation, wind speed, visibility, humidity, temperature, cloud cover) are obtained from Météo-France and assigned to each link using a nearest-station approach.

\textbf{Volume measurements.}
Hourly traffic counts are retrieved from the AVATAR platform (CEREMA) and mapped to directed HERE links for the Lyon and Nantes networks. Dataset statistics are summarized in Table~\ref{tab:datasets}, where we report the effective number of available daily volume profiles (\#Samples), accounting for incomplete or missing measurements over the considered time spans. 

\begin{table}[htbp]
\centering
\small
\setlength{\tabcolsep}{4pt}
\renewcommand{\arraystretch}{1.1}
\begin{tabular}{lccc}
\toprule
City & \#Sensors & Timespan & \#Samples \\
\midrule
Lyon    & 1025 & 01/01/2025 -- 14/12/2025 & 280576 \\
Nantes  & 130  & 17/11/2025 -- 30/11/2025 & 2213 \\
\bottomrule
\end{tabular}

\caption{Summary of the gathered volume measurements.}
\label{tab:datasets}
\vspace{-0.7cm}

\end{table}

\subsection{Experimental Setting}

\textbf{Training and evaluation protocol.}
We adopt nested cross-validation with ensembling on the Lyon dataset. Five outer folds (80/20 split) are used for generalization assessment, with outer test sets fully held out from model selection. Within each outer training fold, three inner validation splits (12.5\%) are used for hyperparameter tuning and early stopping. The three best checkpoints from each inner split are retained, yielding $9$ models per outer fold, whose predictions are averaged. For inter-network evaluation, each outer-fold ensemble is also directly applied to all Nantes sensors without retraining. All results are reported as mean$\pm$std over the five outer folds.

\textbf{Model and optimization details.}
Experiments are conducted on a workstation with an NVIDIA GeForce RTX 4070 GPU. Hyperparameters are manually tuned. Input speed and target volume profiles use a temporal horizon of $T=24$ (hourly resolution). The model includes $C_{\text{static}}=10$ static and $C_{\text{temp}}=12$ temporal features per timestep, over $\mathcal{D}=348$ working days. Both branches use $L_{\text{static}}=L_{\text{temp}}=2$ layers with hidden dimensions $C_{\text{sh}}=C_{\text{th}}=64$. The capacity head consists of three linear layers with hidden dimension 64. The three MLPs of the regime-aware gating each use two linear layers. Transformer encoders have model dimension 64, feedforward dimension 128, and 4 attention heads with sinusoidal positional encoding. $\mathrm{ReLU}$ activations are used throughout, except $\tanh$ in the capacity head. Dropout (0.1--0.6) is applied after each layer. Batch size is 256.

The model is trained by minimizing the composite loss
\[
\mathcal{L} = \mathcal{L}_{\text{vol}} + \lambda_{\text{cap}} \, \mathcal{L}_{\text{cap}},
\]
where $\mathcal{L}_{\text{cap}}$ denotes the capacity-related loss defined previously and $\lambda_{\text{cap}}$ is a weighting coefficient, fixed at 0.05 for all experiments. The volume loss is defined as a Mean Absolute Error (MAE) over supervised link--day pairs:

\[
\mathcal{L}_{\text{vol}} 
= 
\frac{1}{|\Omega|}
\sum_{(\ell,d)\in\Omega}
\left(
\frac{1}{T}
\sum_{t=1}^{T}
\left| \hat{q}_{\ell,d,t} - q_{\ell,d,t} \right|
\right).
\]

The MAE is preferred over squared-error alternatives as it provides a more balanced learning signal across the full range of traffic volumes, preventing large flows from disproportionately dominating the optimization process while still preserving sensitivity to high-demand conditions.

\textbf{Baselines and ablations.}
To quantify the contribution of each structural component for the model spatial generalization abilities, we conduct a systematic ablation analysis :
\begin{itemize}
    \item (1) removes the capacity–utilization decomposition in the prediction head and replaces it with a standard MLP that directly estimates traffic volume, without incorporating any traffic-theoretic structural constraints.
    \item (2) removes explicit supervision of the capacity head, relying exclusively on indirect optimization through the volume prediction loss.
    \item (3) removes the mixture-of-regimes gating mechanism and models traffic dynamics using a single utilization ratio, thereby unifying pre-capacity and post-capacity regimes under a single representation.
\end{itemize}

We also compare the proposed framework against TrafficFlowNet (TFN) \cite{hammoumi2025}, a multimodal transformer-based model representative of the inductive local mapping approaches covered in our work. To ensure a fair comparison, we incorporate the same temporal encoding, including calendar and weather variables, where it is concatenated with the speed attributes before being processed by the model.

\textbf{Evaluation Metrics.}
Hourly link-level volume predictive performance is evaluated using the RMSE, MAPE, and GEH statistics \cite{b25}, along with the sample proportion for which $\mathrm{GEH} > 5$, aggregated over all instrumented links and time intervals. All resulting metrics are shown in Table~\ref{tab:flow_res}.

\begin{table}[htbp]
\centering
\caption{Volume estimation performance in Lyon (nested CV mean$\pm$std) and Nantes (transfer without retraining).}
\label{tab:flow_res}
\renewcommand{\arraystretch}{1.1}

\textbf{Lyon: intra-network generalization}

\begin{tabular*}{\columnwidth}{@{\extracolsep{\fill}}l|cccc}
\toprule
Model & RMSE & MAPE & GEH & \%GEH$>$5 \\
\midrule
\textbf{Our Model} 
& 227.71$\pm$38.70 & 49.54$\pm$4.08 & 6.47$\pm$0.24 & 47.4$\pm$1.4 \\

\midrule
\multicolumn{5}{@{}l}{\textit{Ablations}} \\

(1) & 240.28$\pm$33.91 & 61.29$\pm$6.31 & 6.85$\pm$0.32 & 50.3$\pm$1.7 \\
(2) & 234.40$\pm$38.90 & 56.23$\pm$6.08 & 6.71$\pm$0.31 & 48.9$\pm$1.6 \\
(3) & 235.50$\pm$43.24 & 54.50$\pm$5.67 & 6.58$\pm$0.31 & 48.2$\pm$1.9 \\

\midrule
\textbf{TFN} 
& 260.39$\pm$38.23 & 65.55$\pm$6.77 & 6.94$\pm$0.23 & 51.0$\pm$2.0 \\

\bottomrule
\end{tabular*}

\vspace{1em}

\textbf{Nantes: inter-network generalization}

\begin{tabular*}{\columnwidth}{@{\extracolsep{\fill}}l|cccc}
\toprule
Model & RMSE & MAPE & GEH & \%GEH$>$5 \\
\midrule
\textbf{Our Model} 
& 176.90$\pm$36.89 & 52.58$\pm$4.55 & 6.52$\pm$0.23 & 48.12$\pm$1.7 \\

\midrule
\multicolumn{5}{@{}l}{\textit{Ablations}} \\

(1) & 205.02$\pm$40.31 & 60.41$\pm$4.86 & 6.92$\pm$0.28 & 50.88$\pm$1.4 \\
(2) & 187.65$\pm$37.74 & 58.69$\pm$5.01 & 6.76$\pm$0.29 & 49.54$\pm$1.3 \\
(3) & 188.70$\pm$41.33 & 55.14$\pm$4.97 & 6.71$\pm$0.27 & 48.34$\pm$1.5 \\

\midrule
\textbf{TFN} 
& 217.37$\pm$39.21 & 65.37$\pm$5.12 & 7.01$\pm$0.25 & 51.65$\pm$1.6 \\

\bottomrule
\end{tabular*}
\vspace{-0.4cm}
\end{table}

\section{Results}

\begin{figure}[b]
    \vspace{-0.4cm}
    \centering
    \includegraphics[width=0.47\columnwidth,trim=0.4cm 0.3cm 0cm 1.5cm,clip]{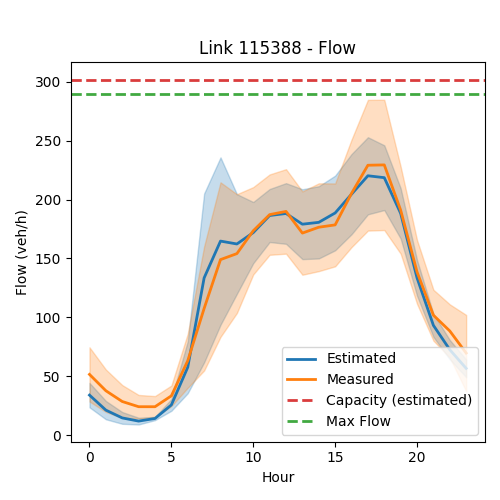}
    \includegraphics[width=0.47\columnwidth,trim=0.4cm 0.3cm 0cm 1.5cm,clip]{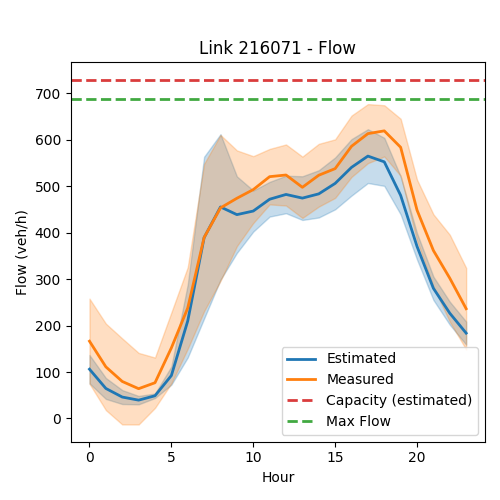}
    \includegraphics[width=0.47\columnwidth,trim=0.4cm 0.3cm 0cm 1.5cm,clip]{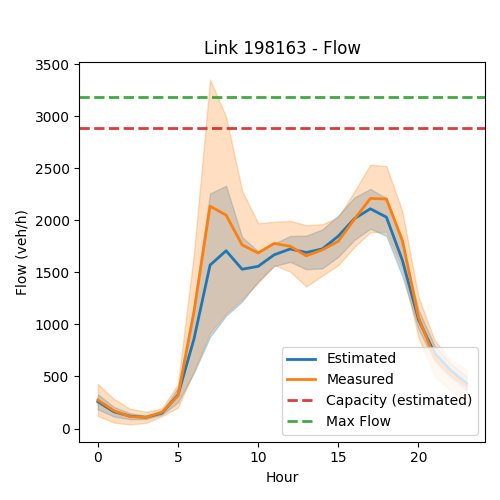}
    \includegraphics[width=0.47\columnwidth,trim=0.4cm 0.3cm 0cm 1.5cm,clip]{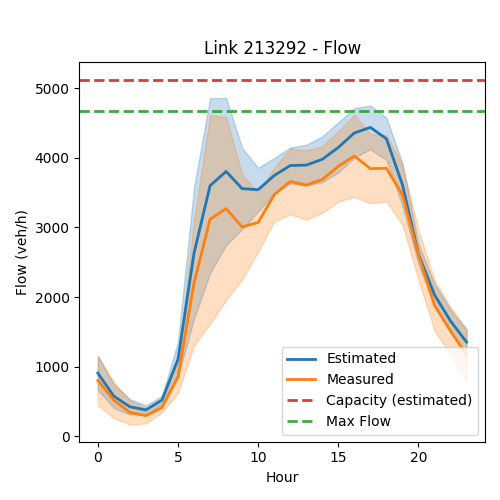}
    \vspace{-0.2cm}
    \caption{Examples of volume estimation on unseen links in the Lyon dataset. For each link, the average estimated (blue) and measured (orange) daily volume profiles over the entire observation period are displayed, with shaded envelopes indicating their variability across the full time span of the dataset.}
    \label{fig:exemples}
\end{figure}

\textbf{Intra-network (Lyon).}
Our capacity-aware model outperforms all ablations and TFN across all metrics. Compared to the direct MLP head (Ablation~1), the capacity-utilization decomposition significantly reduces RMSE and GEH, confirming the benefit of separating structural scale from temporal demand. Removing capacity supervision (Ablation~2) or the regime-aware gating (Ablation~3) also degrades performance, highlighting the importance of both structural anchoring and non-monotonic regime modeling. Figure \ref{fig:exemples} illustrates examples of accurately estimated traffic volume profiles on four previously unseen links in Lyon.

\textbf{Inter-network (Nantes).}
In the cross-network OOD setting, where the model trained on Lyon is directly applied to Nantes without retraining, the same ranking is observed. Our method remains the most robust under spatial distribution shift, indicating that embedding traffic-theoretic constraints improves cross-city transferability.

\begin{figure}[htbp]
\centering

\includegraphics[
    width=\columnwidth,
    trim=0cm 0.5cm 0cm 3cm,
    clip
]{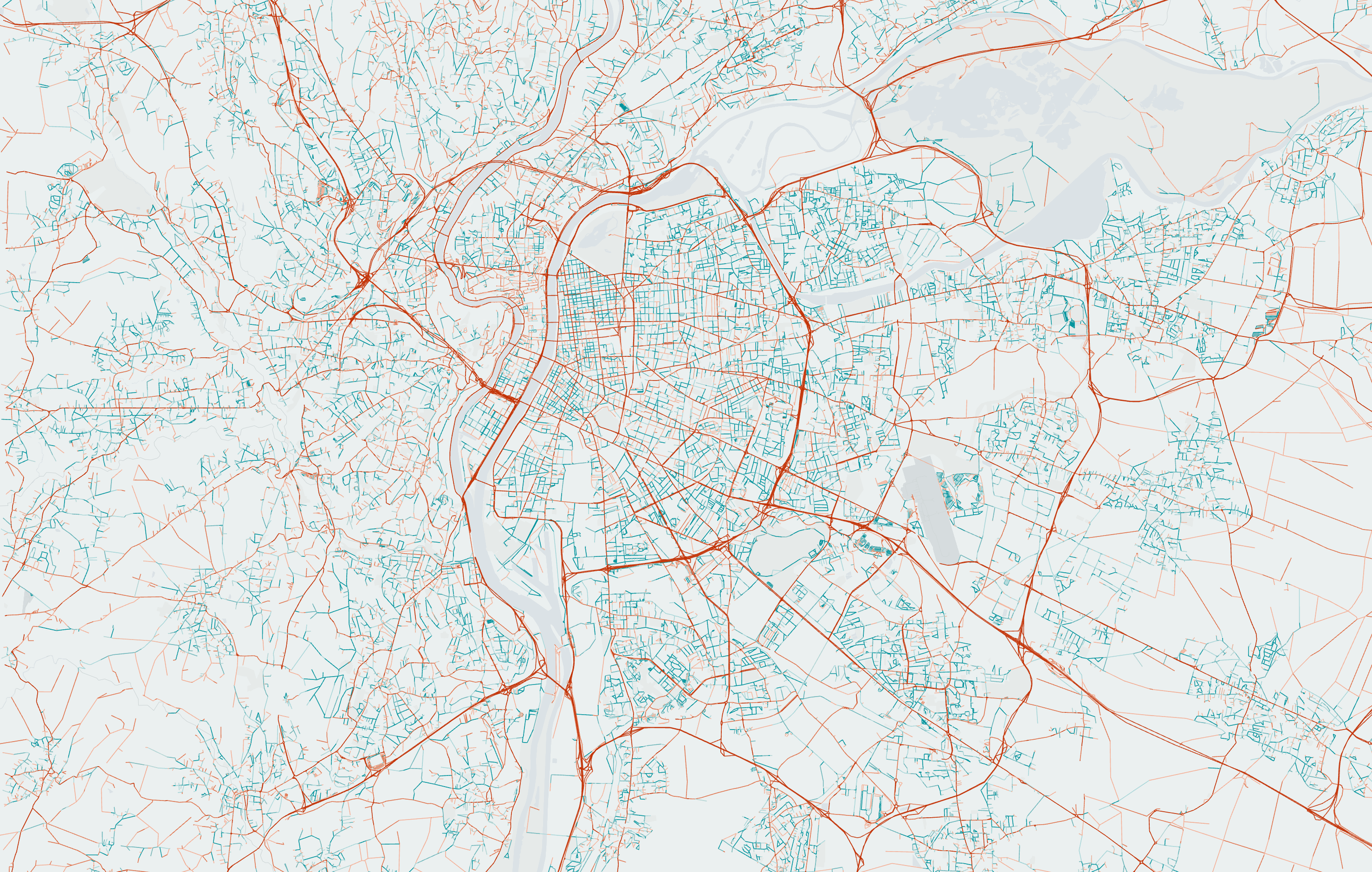}

\vspace{1.5em} 

\includegraphics[
    width=\columnwidth,
    trim=0cm 4cm 0cm 8cm,
    clip
]{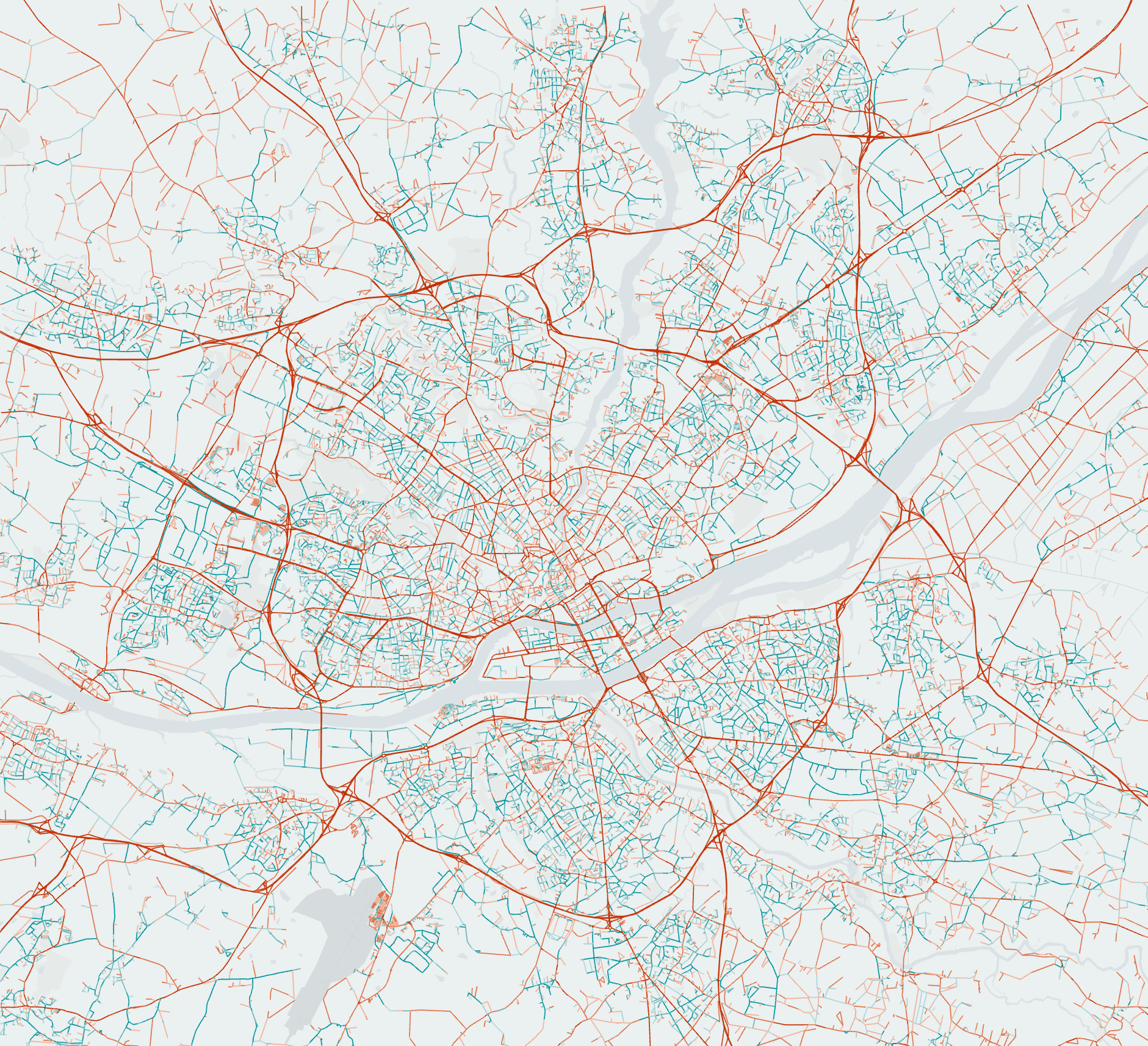}

\caption{Network-wide traffic volume estimates for Lyon (top) and Nantes (bottom) on 18/11/2025 at 9:00 AM. Links are colored according to volume quantiles, where red denotes the highest values and blue the lowest.}
\label{fig:full_network}
\vspace{-0.6em}
\end{figure}

\textbf{Network-wide deployment.}
Figure~\ref{fig:full_network} presents full-network volume predictions for Lyon and Nantes on 18/11/2025 at 9:00 AM. The learned local mapping is applied to all links, yielding spatially coherent hourly flow fields without requiring traffic counts at inference time.

\section{Conclusion}

We proposed a capacity-aware deep learning framework for link-level traffic volume estimation under spatial out-of-distribution generalization. By decomposing volume into a supervised structural capacity and a bounded, regime-aware utilization ratio, the model embeds traffic-theoretic constraints directly into the learning process. Results show consistent improvements in both intra- and inter-network generalization, with gains in average performance metrics and a reduction of large GEH discrepancies. Although the formulation is local at the link level, it provides a scalable and transferable solution for network-wide deployment. Future work could incorporate network-level consistency constraints (e.g., flow conservation) and richer structural priors, as well as leverage more diverse training data to further strengthen OOD generalization.

\section{Acknowledgement}
This work is funded by the French National Research Agency as part of the Mob Sci-Dat Factory project (ANR-23-PEMO-0004) under the France 2030 program.

\end{document}